\documentclass[10pt]{article} 

\usepackage[accepted]{rlj} 

\usepackage{amssymb}            
\usepackage{mathtools}          
\usepackage{mathrsfs}           
\usepackage{graphicx}           
\usepackage{subcaption}         
\usepackage[space]{grffile}     
\usepackage{url}                
\usepackage{lipsum}             

\usepackage[whole]{bxcjkjatype}
\usepackage{cleveref}
\crefname{figure}{Figure}{Figure}
\crefname{table}{Table}{Table}
\crefname{algorithm}{Algorithm}{Algorithm}
\crefname{equation}{Eq.}{Eq.}

\usepackage{bbm}
\usepackage{xspace}
\usepackage{algpseudocode}
\usepackage{algorithm}
\usepackage{wrapfig}
\usepackage{lipsum}
\usepackage{booktabs, multirow} %
\usepackage{soul}%
\usepackage{float}
\usepackage[english]{babel}
\usepackage{amsthm}
\usepackage{tablefootnote}
\usepackage[flushleft]{threeparttable}

\title{Offline Reinforcement Learning with Wasserstein Regularization via Optimal Transport Maps}

\setrunningtitle{Offline Reinforcement Learning with Wasserstein Regularization via Optimal Transport Maps}

\author{Motoki Omura\textsuperscript{1}, Yusuke Mukuta\textsuperscript{1,2}, Kazuki Ota\textsuperscript{1}, Takayuki Osa\textsuperscript{2}, Tatsuya Harada\textsuperscript{1,2}}

\emails{ \{omura,mukuta,ota\}@mi.t.u-tokyo.ac.jp, takayuki.osa@riken.jp, harada@mi.t.u-tokyo.ac.jp}

\affiliations{
$^{1}$\textbf{The University of Tokyo}\\
$^{2}$\textbf{RIKEN}\\
}

\contribution{
    We introduce a novel regularization method with the Wasserstein distance via optimal transport maps for offline RL, eliminating the need for adversarial training and a discriminator through ICNNs.
    }
    {
    \citet{wu2019brac,asadulaev2024rethinking_ot} performed regularization using the Wasserstein distance in offline reinforcement learning through adversarial learning with a discriminator. \citet{makkuva2020wass_otmap_incc,korotin2021wass_wo_minimax,korotin2021wass2dist,mokrov2021large} modeled the Wasserstein distance in a discriminator-free manner using ICNNs in a non-RL domain.
    }
\contribution{
    We evaluate our proposed method on the D4RL benchmark dataset and find that it achieves performance comparable to or even surpassing that of widely used methods. Additionally, by comparing it with an adversarial training-based approach, we show that our discriminator-free method incorporates Wasserstein distance regularization more effectively for these tasks.
    }
    {
    We compared our method with \citet{kostrikov2022iql}, which serves as a component of our approach, and a \citet{wu2019brac}-based method that performs regularization using the Wasserstein distance via discriminator-based adversarial learning. By keeping the value function learning consistent across these existing methods and the proposed method, we fairly evaluated the effect of our proposed regularization on the policy.
    }


\summary{
Offline reinforcement learning (RL) aims to learn an optimal policy from a static dataset, making it particularly valuable in scenarios where data collection is costly, such as robotics. A major challenge in offline RL is distributional shift, where the learned policy deviates from the dataset distribution, potentially leading to unreliable out-of-distribution actions. To mitigate this issue, regularization techniques have been employed. While many existing methods utilize density ratio-based measures, such as the $f$-divergence, for regularization, we propose an approach that utilizes the Wasserstein distance, which is robust to out-of-distribution data and captures the similarity between actions. Our method employs input-convex neural networks (ICNNs) to model optimal transport maps, enabling the computation of the Wasserstein distance in a discriminator-free manner, thereby avoiding adversarial training and ensuring stable learning. Our approach demonstrates comparable or superior performance to widely used existing methods on the D4RL benchmark dataset. The code is available at \url{https://github.com/motokiomura/Q-DOT}. 
}

\begin{document}

\maketitle  

\begin{abstract}
Offline reinforcement learning (RL) aims to learn an optimal policy from a static dataset, making it particularly valuable in scenarios where data collection is costly, such as robotics. A major challenge in offline RL is distributional shift, where the learned policy deviates from the dataset distribution, potentially leading to unreliable out-of-distribution actions. To mitigate this issue, regularization techniques have been employed. While many existing methods utilize density ratio-based measures, such as the $f$-divergence, for regularization, we propose an approach that utilizes the Wasserstein distance, which is robust to out-of-distribution data and captures the similarity between actions. Our method employs input-convex neural networks (ICNNs) to model optimal transport maps, enabling the computation of the Wasserstein distance in a discriminator-free manner, thereby avoiding adversarial training and ensuring stable learning. Our approach demonstrates comparable or superior performance to widely used existing methods on the D4RL benchmark dataset. The code is available at \url{https://github.com/motokiomura/Q-DOT}. 
\end{abstract}


\section{Introduction}
In offline reinforcement learning (RL), learning is conducted solely using a pre-collected dataset to maximize return. When the learned policy deviates from the behavior policy of the dataset, issues such as overestimation of values in unseen states and actions arise \citep{levine2020rltutorial}. Preventing such divergence remains a central challenge in offline RL. Prior studies introduced regularization methods to mitigate distributional shift, including those based on the $f$-divergence \citep{wu2019brac,garg2023extreme,sikchi2024dual}.

Regularization measures based on the density ratio of distributions such as $f$-divergence can become unstable when the supports of the distributions do not overlap, and these measures do not consider the similarity between variables.
Thus, we employ the Wasserstein distance as a regularization term, as it is robust to out-of-distribution data and can incorporate the metric of the variable space. 
When we apply the Wasserstein distance to RL, we can take into consideration the distances in the continuous action space and can handle out-of-distribution actions.



The Wasserstein distance between probability distributions \( P \) and \( Q \) is defined as the infimum of the expected value of the distance between corresponding samples over all possible couplings of \( P \) and \( Q \). For the 2-Wasserstein distance, if \( P \) is absolutely continuous with respect to the Lebesgue measure, there exists a convex function \( \psi \) whose gradient \( \nabla \psi \) acts as the unique optimal transport map from \( P \) to \( Q \). In this setting, the coupling induced by \( P \) and the mapping \( \nabla \psi \) is the optimal coupling \citep{brenier1991polar}.

Here, we consider the case where \( P \) is fixed, and \( Q \) is optimized within an objective function that includes \( W^2_2(P,Q) \). When \( Q \) is directly modeled using a generator and the Wasserstein distance is computed with a discriminator, as in WGAN \citep{arjovsky2017wgan}, additional instability occurs due to adversarial training. Moreover, since controlling the Lipschitz constant of the discriminator is inherently difficult, accurately computing the Wasserstein distance is challenging. To address this issue, we propose optimizing a convex function \( \psi \) in place of \( Q \), based on Brenier's theorem \citep{brenier1991polar}. Since \( \psi \) is learned by minimizing the \( L^2 \) distance instead of using adversarial training, the learning process is relatively stable. Furthermore, as long as \( \psi \) remains convex, it is guaranteed to approximate the Wasserstein distance between \( P \) and some distribution \( Q \), ensuring that the exact Wasserstein distance is consistently computed, even during training.

We apply this approach to policy regularization in offline RL by applying Wasserstein distance regularization to the policy. 
We train the target policy as a transport mapping from the dataset policy $\pi^\mathcal{D}$. In other words, actions sampled from $\pi^\mathcal{D}$ are mapped and treated as actions from the learned policy. By using the gradient of a convex function for this transport map, we aim to learn a mapping that corresponds to the optimal transport map under the 2-Wasserstein distance $W^{2}_{2}(\pi, \pi^\mathcal{D})$.

We employ input-convex neural networks (ICNNs) \citep{amos2017icnn} as the parameterized convex function. By integrating this policy learning method with existing in-sample value function learning methods, we propose a simple Wasserstein regularized algorithm that only requires additional ICNN training. We refer to our approach as \textit{Q-learning regularized by Direct Optimal Transport modeling} (Q-DOT) and evaluate its performance through experiments.

We conduct experiments using the D4RL benchmark dataset \citep{fu2020d4rl} and compare our method with widely used existing approaches. The results demonstrate that our proposed method achieves performance comparable or superior to existing methods. Furthermore, we compare our method with adversarial training-based Wasserstein distance regularization methods that use a discriminator, confirming that our discriminator-free approach is more stable and effective.

Our study makes the following key contributions:
\begin{itemize}
\item We introduce a novel regularization method with the Wasserstein distance via optimal transport maps for offline RL, eliminating the need for adversarial training and a discriminator through ICNNs.
\item We evaluate our proposed method on the D4RL benchmark dataset and find that it achieves performance comparable to or even surpassing that of widely used methods. Additionally, by comparing it with an adversarial training-based approach, we show that our discriminator-free method incorporates Wasserstein distance regularization more effectively for these tasks.
\end{itemize}

\section{Preliminaries}




\subsection{Reinforcement Learning}

Reinforcement learning (RL) is a framework for sequential decision-making, where an agent interacts with an environment modeled as a Markov decision process (MDP). An MDP is defined by the tuple \((\mathcal{S}, \mathcal{A}, P, r, \gamma, d_0)\), where \(\mathcal{S}\) and \(\mathcal{A}\) are the state and action spaces, \(P(s' | s, a)\) is the transition probability distribution, \(r(s, a)\) is the reward function, \(\gamma \in (0,1)\) is the discount factor and $d_0$ is the probability distribution of initial states. The agent follows a policy \(\pi(a | s)\), which defines a probability distribution over actions given a state, aiming to maximize the expected cumulative reward:  $\mathbb{E} [ \sum_{t=0}^{T} \gamma^t r(s_t, a_t) ]$, where $T$ is a task horizon.

\subsection{Offline RL with Regularization}

Offline RL focuses on learning an optimal policy purely from a fixed dataset \(\mathcal{D} = \{(s, a, r, s')\}\) collected by an unknown behavior policy \(\pi_\mathcal{D}\). Since the learned policy \(\pi\) may select actions outside the support of \(\mathcal{D}\), distributional shift issues arise, causing erroneous value estimates and degraded performance.  

To mitigate distributional shift, regularization techniques are employed to constrain the learned policy. 
Regularization is often applied to the divergence between the learned policy \(\pi\) and the dataset policy \(\pi_\mathcal{D}\) \citep{wu2019brac,xu2023sqleql,garg2023extreme}. 
Similarly, our study incorporates regularization on the policy into the objective function.
The optimization problem with regularization is formulated as follows:
\begin{equation}
\label{eq:obj}
\begin{split}
    \max_\pi \quad & \mathbb{E}_{\tau \sim \pi}[Q(s,a) - \alpha D(\pi(\cdot | s) \| \pi^\mathcal{D}(\cdot | s))) ]
\end{split}
\end{equation}
where \( D \) is a divergence that measures the deviation between distributions, and \( \alpha \) is a hyperparameter that adjusts the strength of regularization. 
In \citet{wu2019brac}, methods using various divergences $D$, such as $f$-divergences, were proposed. Directly computing the regularization term requires access to $\pi^\mathcal{D}$, but in offline RL, this policy is often unavailable and must be recovered from the offline dataset. \citet{wu2019brac} circumvented this issue by using adversarial training based on the dual form. However, adversarial training is generally unstable and requires careful hyperparameter tuning. While adversarial training was also used to apply regularization based on the Wasserstein distance, our reproduction found it to be unstable, failing to learn across many tasks. Therefore, in this work, we propose a method that incorporates Wasserstein distance regularization without relying on adversarial training.


\subsection{Wasserstein Distance}

The Wasserstein distance, particularly the 2-Wasserstein distance, is widely used to measure the discrepancy between two probability distributions.  
Given two distributions \( P \) and \( Q \) on $\mathbb{R}^D$ with finite second order moments, the 2-Wasserstein distance is defined as follows:
\begin{equation}
\begin{split}
W_2^2(P, Q) \coloneqq \min_{\xi \in \Pi(P, Q)} \int_{\mathbb{R}^D \times \mathbb{R}^D} \| x - y \|_2^2 \, d\xi(x, y),
\end{split}
\end{equation}
where \( \Pi(P, Q) \) denotes the set of all joint distributions whose marginals are \( P \) and \( Q \).  
The Wasserstein distance captures the geometric discrepancy between probability distributions. Unlike density-ratio-based measures such as the KL divergence, which can diverge when the supports of the distributions do not overlap, the Wasserstein distance is less prone to divergence and serves as a robust measure for out-of-distribution data.

\cite{brenier1991polar} showed that if \(P\) is absolutely continuous with respect to the Lebesgue measure, 
there exists a convex function \( \psi: \mathbb{R}^D \to \mathbb{R} \cup \{\infty\} \) whose gradient \( \nabla \psi:\mathbb{R}^D \rightarrow \mathbb{R}^D \) serves as the unique optimal transport map from \( P \) to \( Q \).
Consequently, the unique optimal transport plan is $\xi^* = [\text{id}_{\mathbb{R}^D}, T^*] \sharp P,$
with \(T^* = \nabla \psi \). Here, for any measurable mapping \(T: \mathbb{R}^D \to \mathbb{R}^D\), \(T \sharp P\) denotes the push-forward of \(P\) under \(T\), and \(\text{id}_{\mathbb{R}^D}\) is the identity map on \(\mathbb{R}^D\).
Then, \(Q = \nabla \psi \sharp P\), and the 2-Wasserstein distance can be expressed as:
\begin{equation}
\begin{split}
W_2^2(P, Q) = \int_{\mathbb{R}^D} \| x - \nabla \psi(x) \|_2^2 \, dP(x).
\end{split}
\end{equation}
The convex function \( \psi(x) \) can be modeled using ICNNs \citep{amos2017icnn}, and its gradient can be used as a mapping to compute the Wasserstein distance \citep{makkuva2020wass_otmap_incc,korotin2021wass2dist,korotin2021wass_wo_minimax,mokrov2021large}.

\section{Offline RL with Wasserstein Regularization via Optimal Transport Maps}
In this section, we propose a method for regularization using the Wasserstein distance in offline RL without relying on a discriminator. 
This algorithm involves learning a value function \( Q_\theta(s,a) \) and \( V_\phi(s) \), a function for transport mapping \( \psi_\omega \), and a policy \( \pi_\rho(a \mid s) \). The parameters \( \theta, \phi, \omega, \rho \) represent the respective neural network parameters.

\subsection{Learning the Visitation Distribution and Policy}
We begin by describing the key component of this approach: learning the transport map \( \nabla \psi_\omega \).
From \cref{eq:obj}, the objective function for \( \psi_\omega \) when the regularization measure is the squared 2-Wasserstein distance is given by:
\begin{equation}
\begin{split}
J_\psi(\omega) = \ \mathbb{E}_{s\sim \mathcal{D},a\sim \pi^\psi}[Q_\theta(s,a) - \alpha W^2_2(\pi^\psi(\cdot | s) \| \pi^\mathcal{D}(\cdot | s))],
\end{split}
\label{eq:obj}
\end{equation}
where $\pi^\psi$ denotes the policy obtained by transporting $\pi^\mathcal{D}$ via the transport map $\nabla \psi_\omega$.

To model \( \pi^\psi \), we apply Brenier's theorem. Specifically, we parameterize a convex function \( \psi_\omega \) using an ICNN and define \( \pi^\psi \) as the push-forward from \( \pi^\mathcal{D} \) through the gradient: $\pi^\psi = \nabla \psi_\omega \sharp \pi^\mathcal{D}$, assuming that $\pi^\mathcal{D}$ is absolutely continuous.
This definition means that samples from \( \pi^\psi \) are obtained as the gradient of the convex function \( \nabla \psi_\omega \), where samples \( a \) are drawn from the offline dataset policy \( \pi^\mathcal{D} \). 

Since the policy $\pi^\psi(\cdot \mid s)$ defines a distribution over actions conditioned on state $s$, the action mapping via $\nabla \psi_\omega$ should also depend on $s$. Therefore, in our implementation, the convex function $\psi_\omega$ takes both the state $s$ and the action $a$ as inputs. After computing the gradient of $\psi_\omega$, we use only the gradient with respect to $a$, i.e., $\nabla_a \psi_\omega(s, a)$, as the transport mapping.
Note that even with this approach, $\psi_\omega$ remains convex with respect to $a$, satisfying the conditions of Brenier's theorem.
This enables mappings that are conditioned on the state $s$.

According to Brenier's theorem, the 2-Wasserstein distance between distributions can be represented by a unique mapping given by the gradient of a convex function. Conversely, due to this uniqueness, if such a convex mapping exists, the $l_2$ distance between the original and mapped distributions always corresponds to the 2-Wasserstein distance.
Consequently, the Wasserstein distance can be evaluated as:
\begin{equation}
\begin{split}
W^2_2(\pi_\omega(\cdot | s) \| \pi^\mathcal{D}(\cdot| s)) = \mathbb{E}_{a \sim \pi^\mathcal{D}} \left[\|a - \nabla_a \psi_\omega(s,a)\|_2^2\right],
\end{split}
\end{equation}
Accordingly, the objective function for \( \psi_\omega \) is formulated as:
\begin{equation}
\begin{split}
\label{eq:psi}
J_\psi(\omega) = \mathbb{E}_{(s,a) \sim \mathcal{D}} \Big[ Q_{\hat{\theta}} (s,\nabla_a \psi_\omega(s,a)) - \alpha \| a - \nabla_a \psi_\omega(s,a)\|_2^2 \Big].
\end{split}
\end{equation}
where \( \hat{\theta} \) represents the parameters of the target network.
We use Input Convex Neural Networks (ICNNs) \citep{amos2017icnn} to represent the parameterized convex function $\psi_\omega$. 
ICNNs are structured using layers where each activation is computed as a convex, non-decreasing function of a non-negative weighted sum of previous activations and direct input passthroughs, ensuring convexity with respect to the input.

In this manner, Wasserstein distance regularization is incorporated into learning without requiring a discriminator. 
This modeling approach offers additional benefits. \citet{makkuva2020wass_otmap_incc} reported that while mappings using conventional neural networks, such as those in \citet{arjovsky2017wgan}, are constrained to be continuous, gradient-based modeling allows for the learning of discontinuous mappings. Our method also benefits from this property. However, in general settings, gradient-based methods face challenges, such as the inability to generate more samples than those available in the offline dataset and the tendency to collapse into an identity mapping when return maximization is absent, preventing the generation of new data.
Nevertheless, in offline RL, it is reasonable to use only reliable data transformed from the offline dataset to avoid out-of-distribution issues. Moreover, the hyperparameter $\alpha$ allows for adjustment between behavior cloning, which corresponds to the identity mapping, and RL. In other words, this method enables discriminator-free modeling without significant issues, making it well-suited for offline RL.

Using the above method, we can learn $\pi^\psi$, which is represented as a mapping from actions in the dataset. However, this approach has a limitation: it cannot sample actions for state-action pairs that are not present in the dataset. To address this issue, we additionally learn another policy $\pi_\rho$, from which we can actually sample actions.
To train $\pi_\rho$ from $\pi^\psi$, we utilize **Advantage Weighted Regression (AWR)** \citep{nair2021awac}, a method commonly used in offline RL.
While existing approaches such as \citet{nair2021awac,kostrikov2022iql,garg2023extreme} maximize the log-likelihood of offline dataset state-action pairs weighted by the advantage, our method instead maximizes the log-likelihood of state-action pairs sampled from $ \pi^\psi$ ( $ = \nabla \psi_\omega \sharp \pi^\mathcal{D} $ ), which are obtained as transformed action of offline data through \( \nabla \psi_\omega \). 
Thus, the loss function is formulated as follows:
\begin{equation}
\begin{split}
\label{eq:pi}
L_\pi(\rho) =  - \mathbb{E}_{(s,a) \sim \mathcal{D}} \left[ \exp \left( \beta (Q_\theta(s, \nabla_a \psi_\omega(s,a)) - V_\phi(s)) \right) \log \pi_\rho(\nabla_a \psi_\omega(s,a) | s) \right].
\end{split}
\end{equation}

\subsection{Learning the Value Function}
There are two common approaches to incorporating regularization into policy learning: \textit{value penalty} and \textit{policy regularization} \citep{wu2019brac}. Value penalty methods introduce a penalty term into the value function, whereas policy regularization directly applies a penalty to the policy itself. 
We adopt a policy regularization approach, in which the value function is learned without any penalty terms. Specifically, we employ Implicit Q-Learning (IQL) \citep{kostrikov2022iql}.
IQL enables Q-learning–based value function learning through the expectile regression, allowing for in-sample learning. As a result, it can learn the optimal Q-values without the need for explicit penalty terms, while avoiding out-of-distribution actions.
The loss functions for \( Q \) and \( V \) are formulated as follows:
\begin{equation}
\begin{split}
\label{eq:v}
L_V(\phi) = \mathbb{E}_{(s,a) \sim \mathcal{D}} \left[ L_{\tau}^2 (Q_{\hat{\theta}}(s, a) - V_{\phi}(s)) \right],
\end{split}
\end{equation}
\begin{equation}
\begin{split}
\label{eq:q}
L_Q(\theta) = \mathbb{E}_{(s,a,s') \sim \mathcal{D}} \left[ (r(s, a) + \gamma V_{\phi}(s') - Q_{\theta}(s, a))^2 \right].
\end{split}
\end{equation}
By integrating this value function learning approach with the previously described learning of \( \psi_\omega \) and \( \pi_\rho \), we achieve a policy regularization-based method that incorporates the Wasserstein distance regularization without relying on a discriminator or adversarial training.

We name this method \textit{Q-learning regularized by Direct Optimal Transport modeling} (Q-DOT) and evaluate it through experiments. The corresponding pseudocode is presented in \cref{alg}.

\begin{wrapfigure}{R}{0.5\textwidth}
    \vspace{-2.1em}
    \begin{minipage}{0.5\textwidth}
        \begin{algorithm}[H]
        \caption{Q-DOT}
        \label{alg}
        \begin{algorithmic}[1]
            \State {\bf Input:} Offline dataset $\mathcal{D} = \{(s, a, r, s')\}$
            \State {\bf Initialize:} $Q_{\theta}, V_{\phi}$, $\pi_{\rho}$ and ICNN $\psi_{\omega}$
            \For{each update step}
                \State Sample mini-batch $\{(s, a, r, s')\}$ from $\mathcal{D}$
                \State Update $V_{\phi}$ by minimizing \cref{eq:v}
                \State Update $Q_{\theta}$ by minimizing \cref{eq:q}
                \State Update $\psi_{\omega}$ by maximizing \cref{eq:psi}
                \State Update $\pi_{\rho}$ by minimizing \cref{eq:pi}
            \EndFor
        \end{algorithmic}
        \end{algorithm}
    \end{minipage}
    \vspace{-2em}
\end{wrapfigure}



\section{Experiments}
\subsection{Experimental Setup}
In this section, we evaluate the effectiveness of the proposed method using the D4RL benchmark \citep{fu2020d4rl}.
For comparison, we consider widely used offline RL methods \citep{fujimoto2021td3bc,kostrikov2022iql,NEURIPS2020cql,NEURIPS2021dt}, and refer to the scores reported in \citet{kostrikov2022iql}. In addition, we implement and experiment with an adversarial learning-based method that incorporates regularization using the Wasserstein distance, which we refer to as Adversarial Wasserstein (AdvW).
In AdvW, the policy is updated based on \citet{wu2019brac}, and its objective is defined as follows:
\begin{equation}
\begin{split}
\max_{\pi} \mathbb{E}_{s \sim \mathcal{D},a \sim \pi}  \left[ Q(s, a) 
- \alpha W_1 (\pi(\cdot | s) \| \pi_\mathcal{D}(\cdot | s)) 
\right].
\end{split}
\end{equation}
where \(\pi_\mathcal{D}\) represents the behavior policy used for dataset collection. The Wasserstein regularization is computed using the dual form with a discriminator \( g \), following \citet{arjovsky2017wgan}: $W(p, q) = \sup_{g : \| g \|_{L} \leq 1} \left( \mathbb{E}_{x \sim p} [g(x)] - \mathbb{E}_{x \sim q} [g(x)] \right)$.
The training of both the discriminator and the policy follows the official implementation of \citet{wu2019brac}. The discriminator is trained with a gradient penalty to enforce the Lipschitz constraint.  
For value function training, we observed that on-policy training, as in \citet{wu2019brac}, failed to achieve decent learning in the D4RL benchmark. 
Therefore, we adopted in-sample learning of IQL, similar to our proposed method, resulting in a fair comparison.
The hyperparameter \(\alpha\) in AdvW was tuned using the values \((0.3, 1, 3, 10, 30)\) specified in \citet{wu2019brac}. However, we found that the learned policy outputs sometimes deviated significantly from the dataset actions. To address this, we further tested larger values \((10^2, 15^2, 20^2)\).

The expectile parameter $\tau$ used for value function estimation in both our proposed method and AdvW is a hyperparameter, and we adopted the same values as in \citet{kostrikov2022iql}. 
The implementation of ICNN utilizes \citet{cuturi2022ott}. The network consists of a two-layer fully connected architecture with 256 hidden units per layer, identical to the other actor and critic networks.
Additional implementation details and other hyperparameters are provided in the Supplementary Materials.


\subsection{Results on the D4RL Benchmark}



\begin{table}[t]
    \centering
    \caption{ The average normalized return on D4RL tasks. For our method, Q-DOT, the mean and standard error over six random seeds are reported.}
    \label{tab:results}
    \resizebox{\textwidth}{!}{%
    \begin{tabular}{l||rrrrrrrr}
        Dataset & BC & 10\%BC & DT & TD3+BC & CQL & IQL & AdvW & Q-DOT (Ours) \\
        \hline
        halfcheetah-medium-v2 & 42.6 & 42.5 & 42.6 & \textbf{48.3} & 44.0 & \textbf{47.4} & \textbf{48.6} & \textbf{47.9}$_{\pm 0.1}$ \\
        hopper-medium-v2 & 52.9 & 56.9 & 67.6 & 59.3 & 58.5 & 66.3 & 61.2 & \textbf{76.7}$_{\pm 3.2}$ \\
        walker2d-medium-v2 & 75.3 & 75.0 & 74.0 & \textbf{83.7} & 72.5 & 78.3 & 80.7 & \textbf{83.0}$_{\pm 0.8}$ \\
        halfcheetah-medium-replay-v2 & 36.6 & 40.6 & 36.6 & \textbf{44.6} & \textbf{45.5} & \textbf{44.2} & 44.2 & \textbf{43.7}$_{\pm 0.6}$ \\
        hopper-medium-replay-v2 & 18.1 & 75.9 & 82.7 & 60.9 & 95.0 & 94.7 & 48.1 & \textbf{97.4}$_{\pm 1.2}$ \\
        walker2d-medium-replay-v2 & 26.0 & 62.5 & 66.6 & \textbf{81.8} & 77.2 & 73.9 & 68.9 & 70.7$_{\pm 4.0}$ \\
        halfcheetah-medium-expert-v2 & 55.2 & \textbf{92.9} & 86.8 & 90.7 & \textbf{91.6} & 86.7 & 22.6 & 89.6$_{\pm 1.7}$ \\
        hopper-medium-expert-v2 & 52.5 & \textbf{110.9} & 107.6 & 98.0 & 105.4 & 91.5 & 17.6 & 93.1$_{\pm 13.0}$ \\
        walker2d-medium-expert-v2 & 107.5 & \textbf{109.0} & 108.1 & \textbf{110.1} & \textbf{108.8} & \textbf{109.6} & 92.5 & \textbf{110.3}$_{\pm 0.1}$ \\
        \hline
        locomotion total & 466.7 & 666.2 & 672.6 & 677.4 & 698.5 & 692.4 & 480.1 & \textbf{712.4} \\
        \hline
        antmaze-umaze-v0 & 54.6 & 62.8 & 59.2 & 78.6 & 74.0 & \textbf{87.5} & 83.2 & \textbf{87.8}$_{\pm 1.1}$ \\
        antmaze-umaze-diverse-v0 & 45.6 & 50.2 & 53.0 & 71.4 & \textbf{84.0} & 62.2 & 51.0 & 70.2$_{\pm 3.8}$ \\
        antmaze-medium-play-v0 & 0.0 & 5.4 & 0.0 & 10.6 & 61.2 & \textbf{71.2} & 46.0 & 68.2$_{\pm 1.5}$ \\
        antmaze-medium-diverse-v0 & 0.0 & 9.8 & 0.0 & 3.0 & 53.7 & \textbf{70.0} & 42.5 & 66.2$_{\pm 5.5}$ \\
        antmaze-large-play-v0 & 0.0 & 0.0 & 0.0 & 0.2 & 15.8 & 39.6 & 12.5 & \textbf{49.0}$_{\pm 4.2}$ \\
        antmaze-large-diverse-v0 & 0.0 & 6.0 & 0.0 & 0.0 & 14.9 & \textbf{47.5} & 8.2 & 40.7$_{\pm 4.9}$\\
        \hline
        antmaze total & 100.2 & 134.2 & 112.2 & 163.8 & 303.6 & 378.0 & 243.3 & \textbf{382.0} \\
        \hline
        kitchen-complete-v0 & \textbf{65.0} & - & - & - & 43.8 & 62.5 & 4.2 & \textbf{64.2}$_{\pm 3.4}$ \\
        kitchen-partial-v0 & 38.0 & - & - & - & 49.8 & 46.3 & 24.6 & \textbf{71.3}$_{\pm 1.3}$ \\
        kitchen-mixed-v0 & \textbf{51.5} & - & - & - & \textbf{51.0} & \textbf{51.0} & 21.7 & 42.9$_{\pm 4.3}$ \\
        \hline
        kitchen total & 154.5 & - & - & - & 144.6 & 159.8 & 50.4 & \textbf{178.3} \\
        \hline
    \end{tabular}%
    }
\end{table}

\cref{tab:results} presents the results on the D4RL benchmark, which consists of the locomotion, antmaze, and kitchen domains. In terms of total return, our proposed method consistently achieved the best or comparable performance across all domains compared to the baselines.  
When analyzing individual datasets, our method attained the highest return on several datasets. Notably, it outperformed the best existing methods by significant margins in hopper-medium-v2 and kitchen-partial-v0, achieving improvements of +10.4 and +21.5, respectively.  
For the Antmaze domain, our method performed similarly to IQL. Since Antmaze requires trajectory stitching \citep{zhuang2024reinformer_stitching}, regularization alone does not substantially enhance stitching ability. 
This suggests that further performance improvements would require incorporating additional techniques beyond regularization.  

On the other hand, AdvW exhibited lower performance across all domains in terms of total score. In particular, it performed poorly on datasets that contain successful trajectories, such as halfcheetah-medium-expert-v2, hopper-medium-expert-v2, and kitchen-complete-v0, where strong regularization is crucial. This was the case even when using large values for \(\alpha\) (e.g., \(10^2, 15^2, 20^2\)).  
These AdvW results are similar to the BRAC results reported in \citet{zhang21bracplus} using a discriminator with $f$-divergence, suggesting that behavior cloning generally becomes more challenging when adversarial learning is involved.

These results suggest that adversarial learning-based regularization via Wasserstein distance is inherently unstable and challenging. In contrast, the discriminator-free training approach of our proposed method demonstrates effectiveness in achieving high scores consistently.

\subsection{Trajectory Quality and Transport Distance}
We analyze the relationship between trajectory quality and action transformations induced by a trained ICNN mapping. Specifically, we visualize the relationship between the cumulative rewards of trajectories in the offline dataset and the average transport-induced distance over actions. The horizontal axis represents the cumulative reward of each trajectory, while the vertical axis indicates the average L2 norm of the difference between actions before and after transport.

\begin{figure}[t]
    \centering
    \begin{minipage}{0.32\textwidth}
        \centering
        \includegraphics[width=\linewidth]{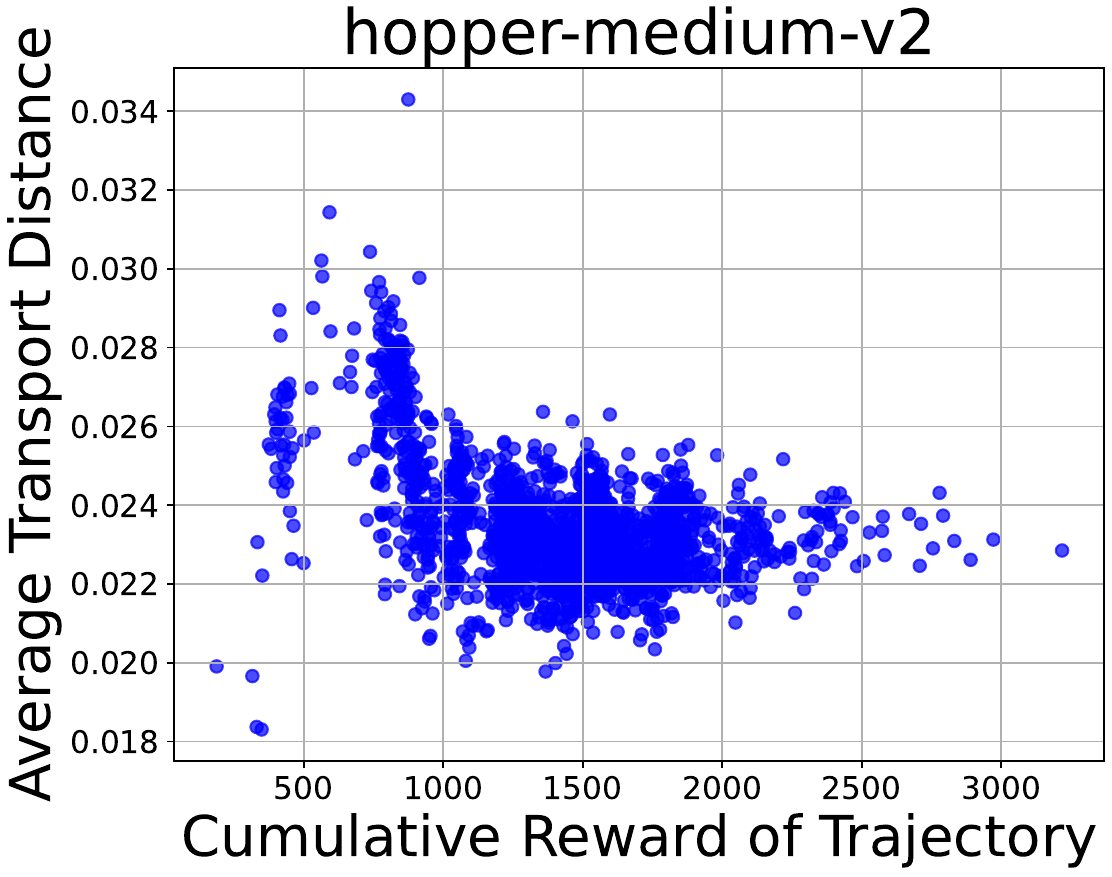}
    \end{minipage}
    \begin{minipage}{0.32\textwidth}
        \centering
        \includegraphics[width=\linewidth]{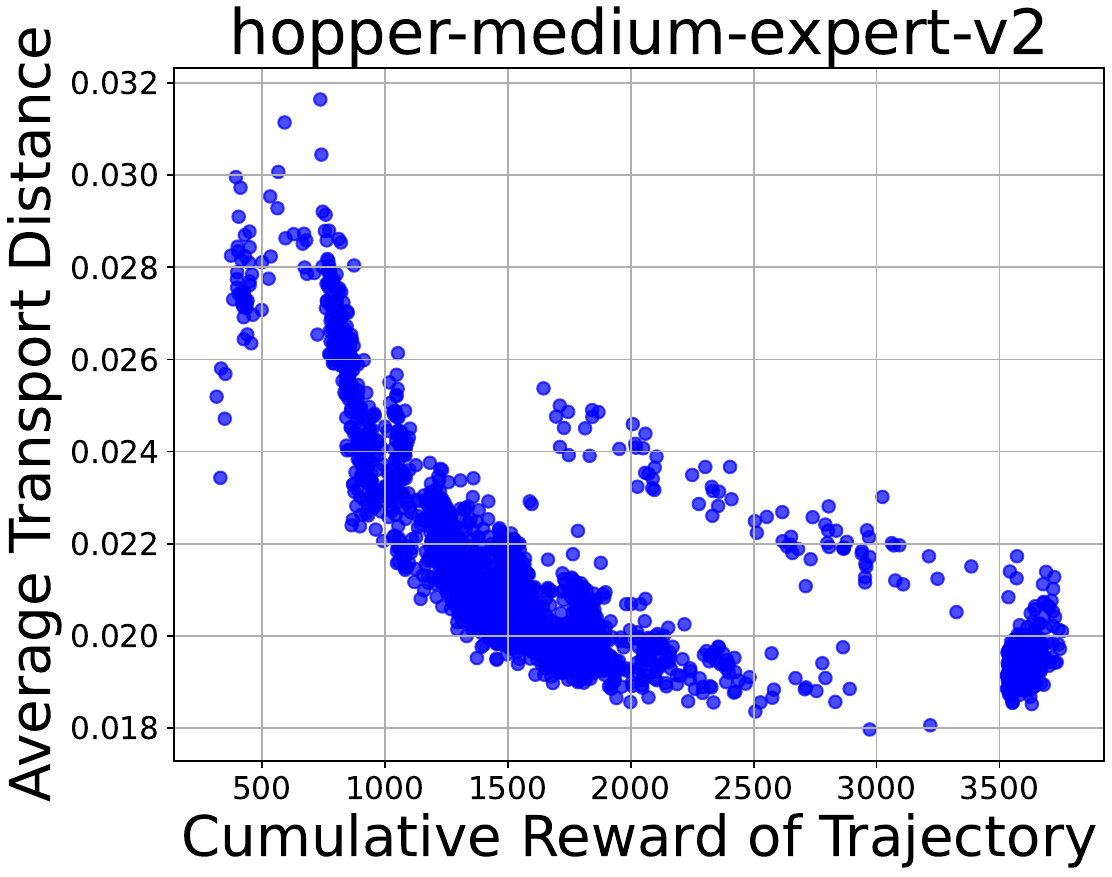}
    \end{minipage}
    \begin{minipage}{0.32\textwidth}
        \centering
        \includegraphics[width=\linewidth]{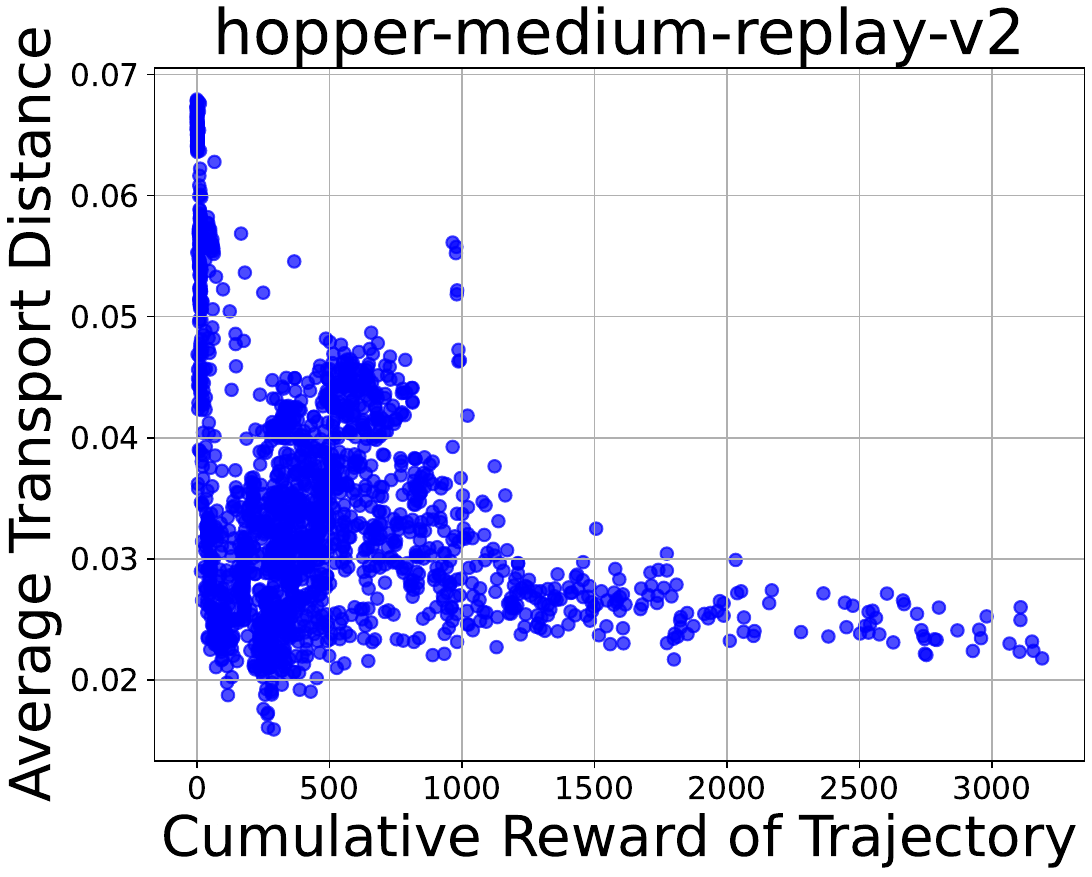}
    \end{minipage}
    \caption{The relationship between trajectory quality and transport-induced distance. The x-axis represents the cumulative reward of each trajectory, while the y-axis shows the average L2 norm of action differences before and after transport.}
    \label{fig:return_dist}
\end{figure}

The results from three tasks show that lower-reward trajectories exhibit greater transport distances for their actions. This suggests that the optimization objective in \cref{fig:return_dist} effectively regularizes action transformations by primarily modifying low-quality trajectories while preserving high-quality ones.  Additional results for other tasks are provided in the Supplementary Material.



\

\section{Related Work}

Offline RL aims to learn policies solely from pre-collected data. A central challenge in this setting is addressing the distributional shift between the state-action distribution of the learned policy and that of the offline dataset. When the distributional shift is large, the value of actions not observed in the offline dataset may be overestimated \citep{levine2020rltutorial}. A particularly simple approach to mitigating this discrepancy is presented in \citep{fujimoto2021td3bc}. \citet{fujimoto2021td3bc} propose a policy regularization method based on TD3, an off-policy technique commonly used in online RL, by incorporating a behavior cloning term into the policy learning process. The behavior cloning term is defined as the squared error between the action output by the learned policy and the action in the dataset. This corresponds to the 2-Wasserstein distance between the dataset policy and the learned policy in cases where the learned policy is deterministic. The promising performance of this simple method suggests the effectiveness of using a notion of action similarity, such as the Wasserstein distance, as a regularization term.

\citet{wu2019brac} experimented with value penalty and policy regularization using various divergence measures. Regularization based on $f$-divergence and the Wasserstein distance was also explored, where optimization was performed using a dual-form discriminator-based approach. However, as demonstrated with AdvW, even when in-sample maximization was incorporated into value function learning, adversarial learning with a discriminator did not perform well on the D4RL dataset, highlighting the necessity of discriminator-free learning.
\citet{asadulaev2024rethinking_ot} proposed a method that extends BRAC with Wasserstein distance regularization. Similar to BRAC, their approach employs adversarial learning using a discriminator. However, their main proposed method is based on \citet{tarasov2023rebrac}, which incorporates a large network and multiple techniques, making a fair comparison challenging.
Their approach, which formulates offline RL as an Optimal Transport problem, could incorporate our ICNN-based modeling, which may be considered a future direction.
\citet{luo2023iqlot} employ optimal transport maps to compute the Wasserstein distance; however, they use them for reward labeling before reinforcement learning, rather than as a regularization technique in offline RL.


Several studies, including \citet{kostrikov2022iql,xu2023sqleql,garg2023extreme,sikchi2024dual}, have proposed in-sample maximization approaches. These methods avoid overestimation caused by out-of-distribution actions by training exclusively with dataset actions, without sampling actions from the learned policy. \citet{kostrikov2022iql} treat the value function as a distribution with inherent action-related randomness and estimates an expectile with $\tau \approx1$ using expectile regression to approximate the optimal value function in an in-sample manner, similar to Q-learning. The policy is learned through Advantage Weighted Regression \citep{nair2021awac}, where behavior cloning is weighted by an advantage function derived from the learned value function, ensuring that regularization is applied only during policy learning. \citet{garg2023extreme,xu2023sqleql} propose algorithms that incorporate regularization terms based on reverse KL divergence and other $f$-divergence measures into the RL objective. 
\citet{sikchi2024dual} employ a regularization term based on the visitation distribution of each policy, following \citet{nachum2020fenchelduality}, where $f$-divergence is used as a measure. Since these in-sample maximization approaches decouple value function learning from policy learning and propose novel methods for value function training, they can be combined with our policy learning method.

Input Convex Neural Networks (ICNNs) \citep{amos2017icnn} are neural networks designed such that their outputs form a convex function with respect to the inputs. Based on Brenier's theorem \citep{brenier1991polar}, the gradient of an ICNN can be utilized as a push-forward map, enabling the modeling of the Wasserstein distance even in high-dimensional data settings \citep{makkuva2020wass_otmap_incc,korotin2021wass2dist,korotin2021wass_wo_minimax,mokrov2021large}. 
The use of an ICNN-based generator for minimizing the Wasserstein distance involves transforming existing data. 
If there is no objective such as return maximization when minimizing the Wasserstein distance using the ICNN-based generator, the transport map simply becomes an identity mapping, rendering it incapable of generating new data and thus meaningless. However, when an objective is introduced, the strength of the regularization term can be adjusted via a hyperparameter $\alpha$, allowing for a gradual increase in the deviation from the identity mapping. These characteristics make Wasserstein regularization using ICNNs especially well-suited to offline RL.
To the best of our knowledge, our proposed approach is the first to introduce discriminator-free Wasserstein distance regularization with ICNNs in RL. 
This method has the potential for further development beyond offline RL, extending to other RL settings.



\section{Conclusion}
In this study, we proposed a novel offline RL method that leverages Wasserstein distance as a regularization technique without requiring adversarial learning with a discriminator. By utilizing the gradient of input-convex neural networks (ICNNs) to model the optimal transport mapping, our approach effectively regularizes the learned policy while maintaining stability and efficiency. 
Through experiments using the D4RL benchmark dataset, we demonstrated that our method performs comparably to or better than established baseline approaches, including adversarial Wasserstein distance regularization methods that rely on a discriminator. These results highlight the effectiveness of our discriminator-free approach in mitigating distributional divergence while ensuring robust policy learning in offline RL settings.
Our findings suggest that Wasserstein distance regularization via ICNN-based optimal transport mapping offers a promising direction for future research in RL.







\subsubsection*{Acknowledgments}
\label{sec:ack}
This research is partially supported by JST Moonshot R\&D Grant Number JPMJPS2011, CREST Grant Number JPMJCR2015 and Basic Research Grant (Super AI) of Institute for AI and Beyond of the University of Tokyo. T.O was supported by JSPS KAKENHI Grant Number JP25K03176. K.O. was supported by JST SPRING, Grant Number JPMJSP2108.


\bibliography{main}
\bibliographystyle{rlj}

\beginSupplementaryMaterials


\section{Experimental Details}

\begin{figure}[h]
    \centering
    \begin{minipage}{0.32\textwidth}
        \centering
        \includegraphics[width=\linewidth]{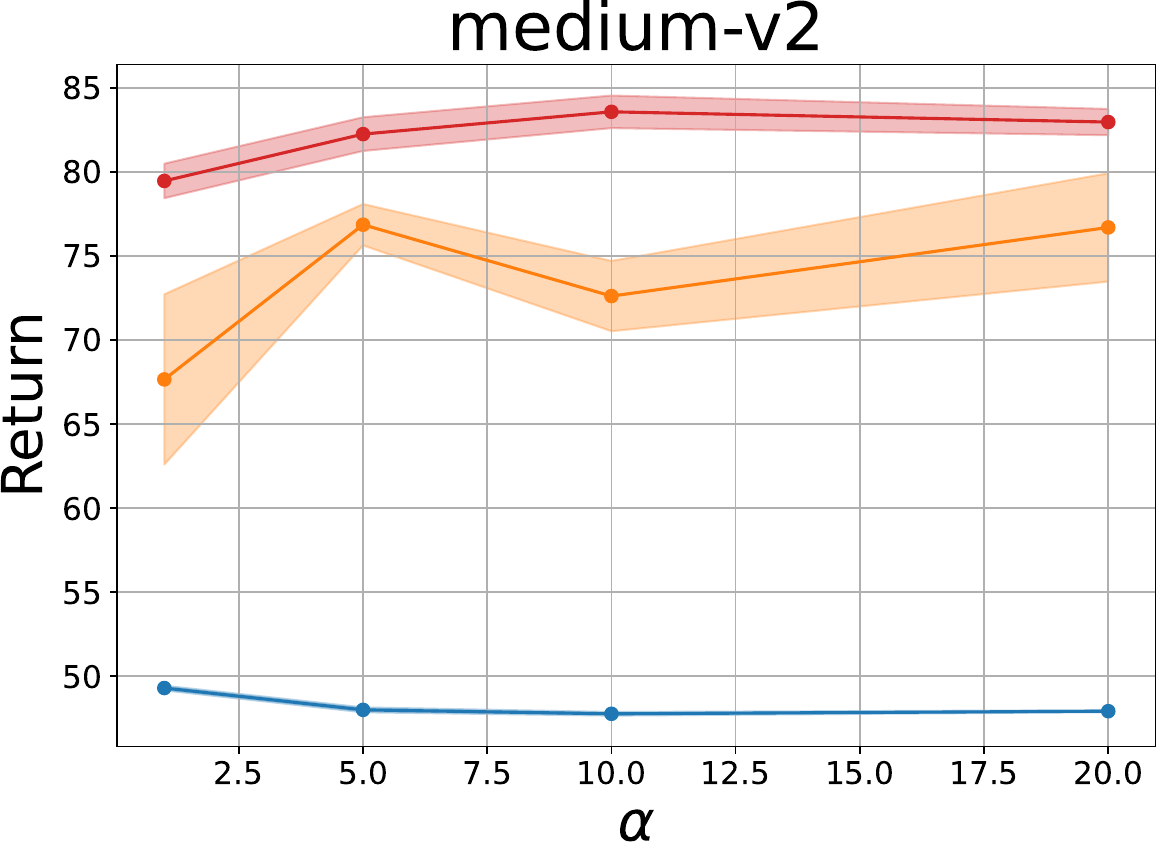}
    \end{minipage}
    \begin{minipage}{0.32\textwidth}
        \centering
        \includegraphics[width=\linewidth]{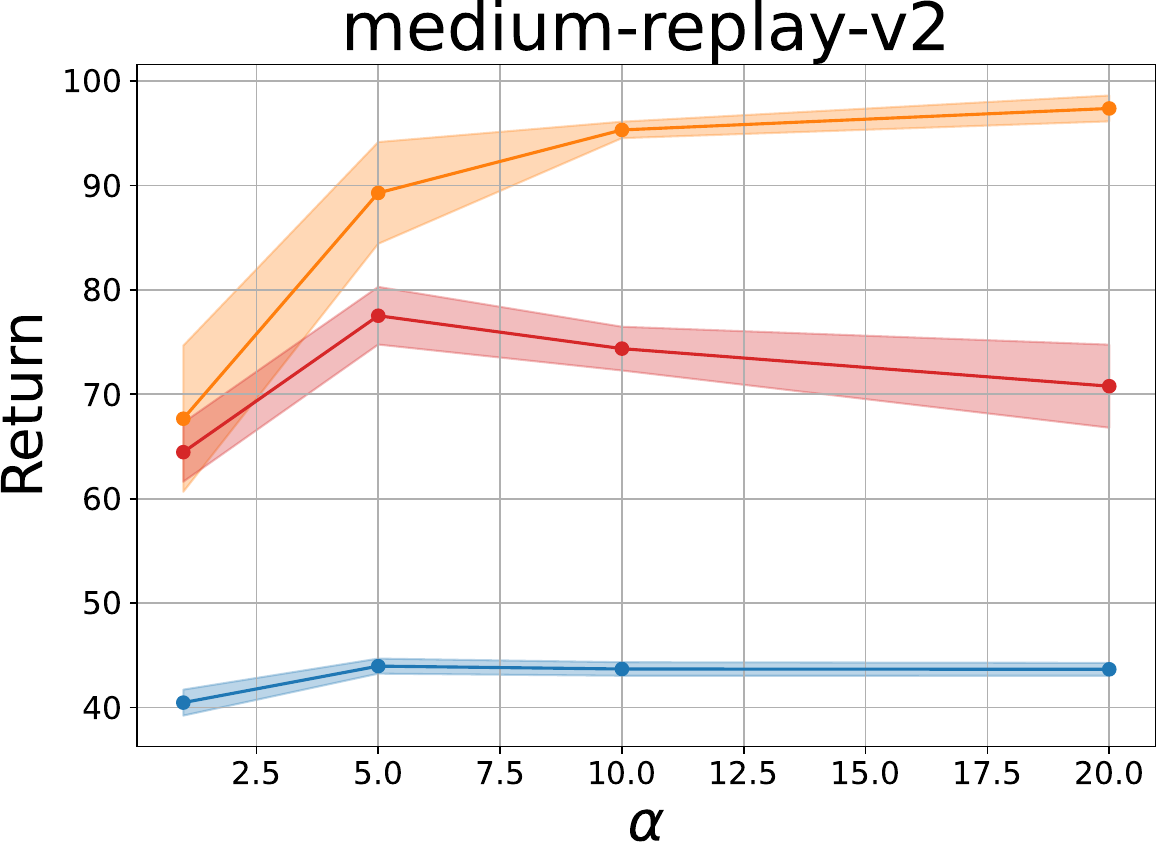}
    \end{minipage}
    \begin{minipage}{0.32\textwidth}
        \centering
        \includegraphics[width=\linewidth]{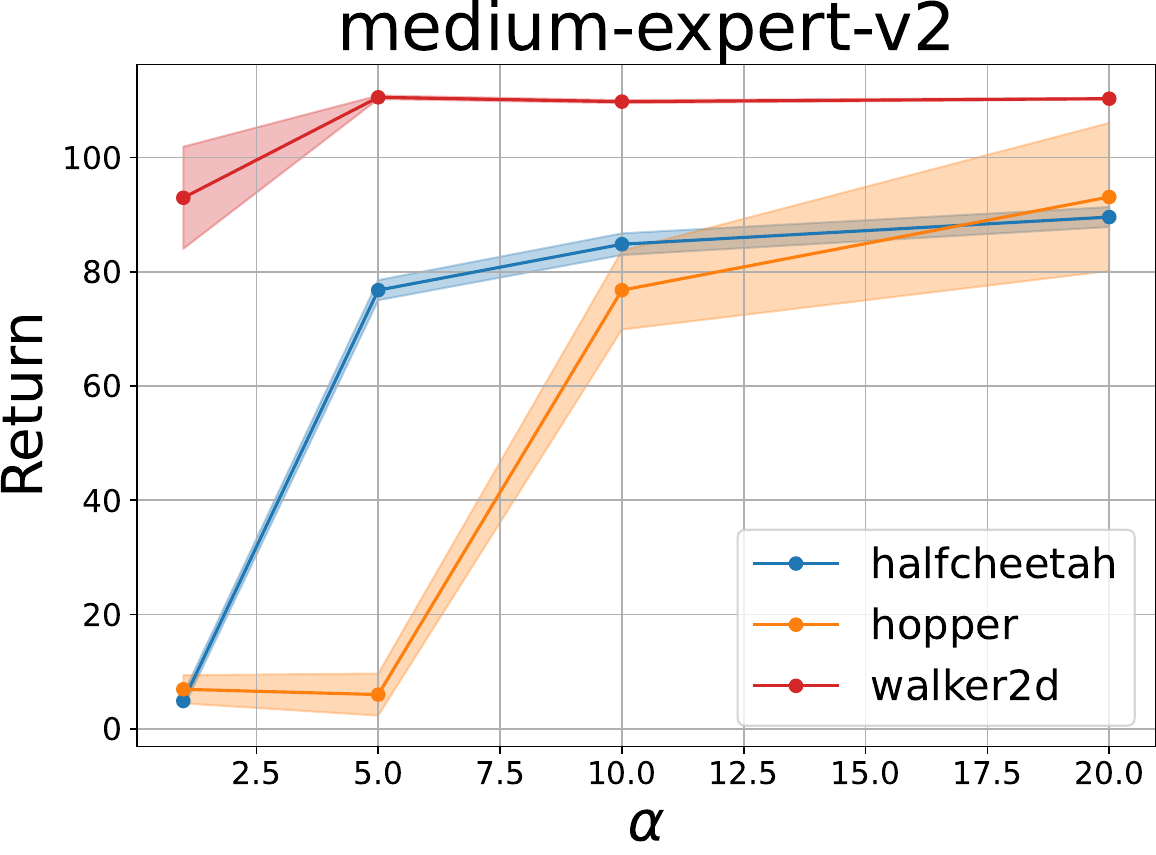}
    \end{minipage}
    \caption{The average returns for each value of $\alpha$ in MuJoCo tasks}
    \label{fig:ablation_alpha}
\end{figure}

In AdvW and Q-DOT, the actor, critic, discriminator (for AdvW), and ICNN (for Q-DOT) are all two-layer MLPs with ReLU activations and 256 hidden units. The learning rate for all updates was set to \(3 \times 10^{-4}\) using the Adam optimizer \citep{kingma2014adam}. The expectile parameter $\tau$ was set to the same value as in IQL: 0.7 for MuJoCo locomotion tasks and Kitchen tasks, and 0.9 for Antmaze tasks.
For AdvW, the parameter \(\alpha\) was selected from the values explored in \citet{wu2019brac} as well as larger values, choosing the optimal one from \((0.3, 1, 3, 10, 30, 10^2, 15^2, 20^2)\). The selected values for MuJoCo locomotion, Antmaze, and Kitchen were 3, 1, and 30, respectively. 
In Q-DOT, \(\alpha\) was selected from \((1, 5, 10, 20, 10^2, 20^2)\), which includes large values, because \(W^2_2\) was often computed as the squared difference of values below 1, resulting in extremely small values. Meanwhile, \(\beta\) was swept over the range \((0.5, 3, 10, 20)\), which is close to the values reported in \citet{kostrikov2022iql}. The selected \((\alpha, \beta)\) pairs for MuJoCo locomotion, Antmaze, and Kitchen were \((20,3)\), \((20,20)\), and \((20^2, 0.5)\), respectively.  
The effect of the parameter $\alpha$ on the return in MuJoCo tasks is shown in \cref{fig:ablation_alpha}.
Other implementation details follow \citet{kostrikov2022iql}. 

\section{Trajectory Quality and Transport Distance}
\begin{figure}[h]
    \centering
    \begin{minipage}{0.32\textwidth}
        \centering
        \includegraphics[width=\linewidth]{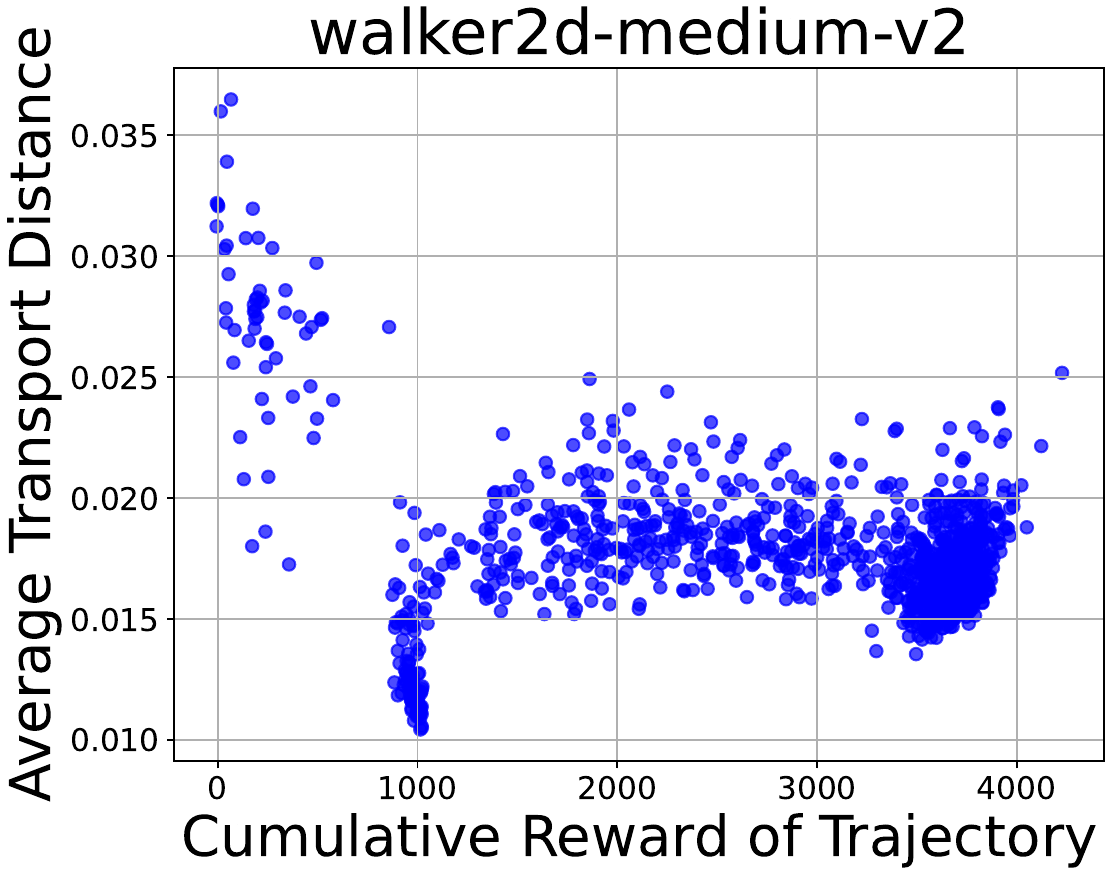}
    \end{minipage}
    \begin{minipage}{0.32\textwidth}
        \centering
        \includegraphics[width=\linewidth]{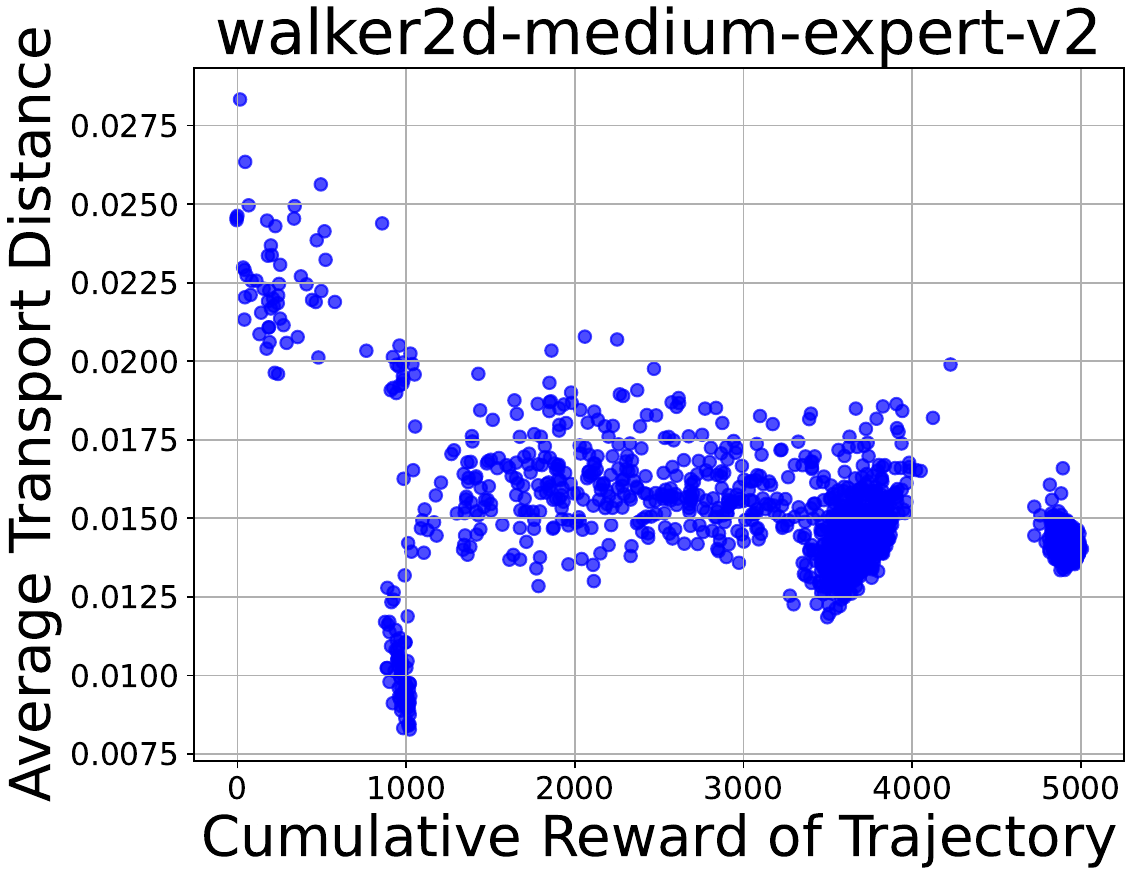}
    \end{minipage}
    \begin{minipage}{0.32\textwidth}
        \centering
        \includegraphics[width=\linewidth]{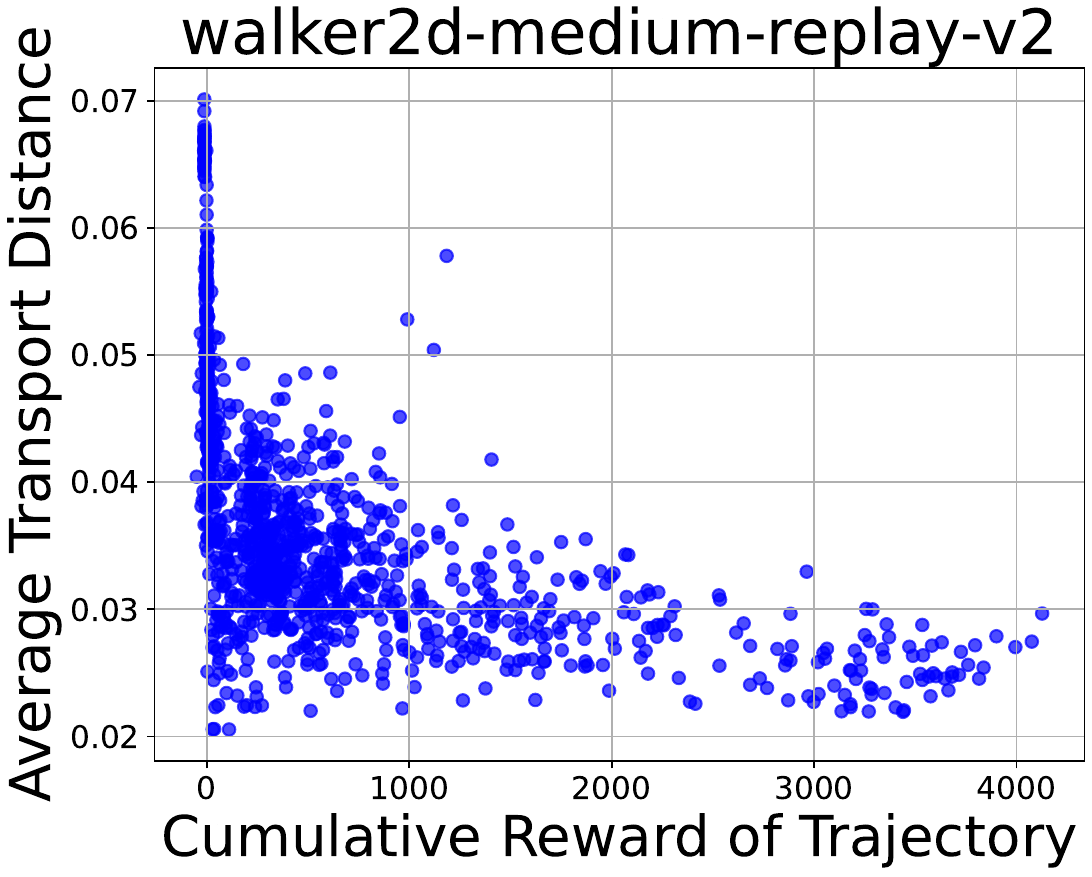}
    \end{minipage}
    
    \begin{minipage}{0.32\textwidth}
        \centering
        \includegraphics[width=\linewidth]{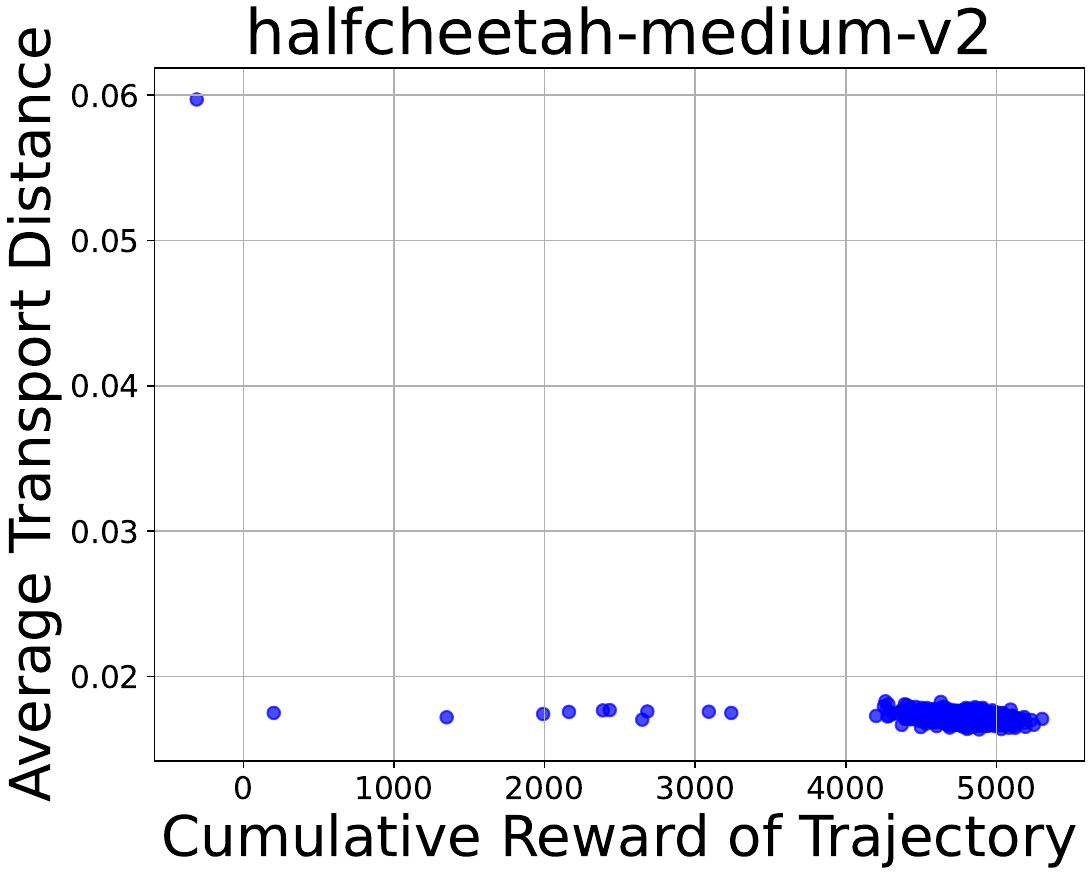}
    \end{minipage}
    \begin{minipage}{0.32\textwidth}
        \centering
        \includegraphics[width=\linewidth]{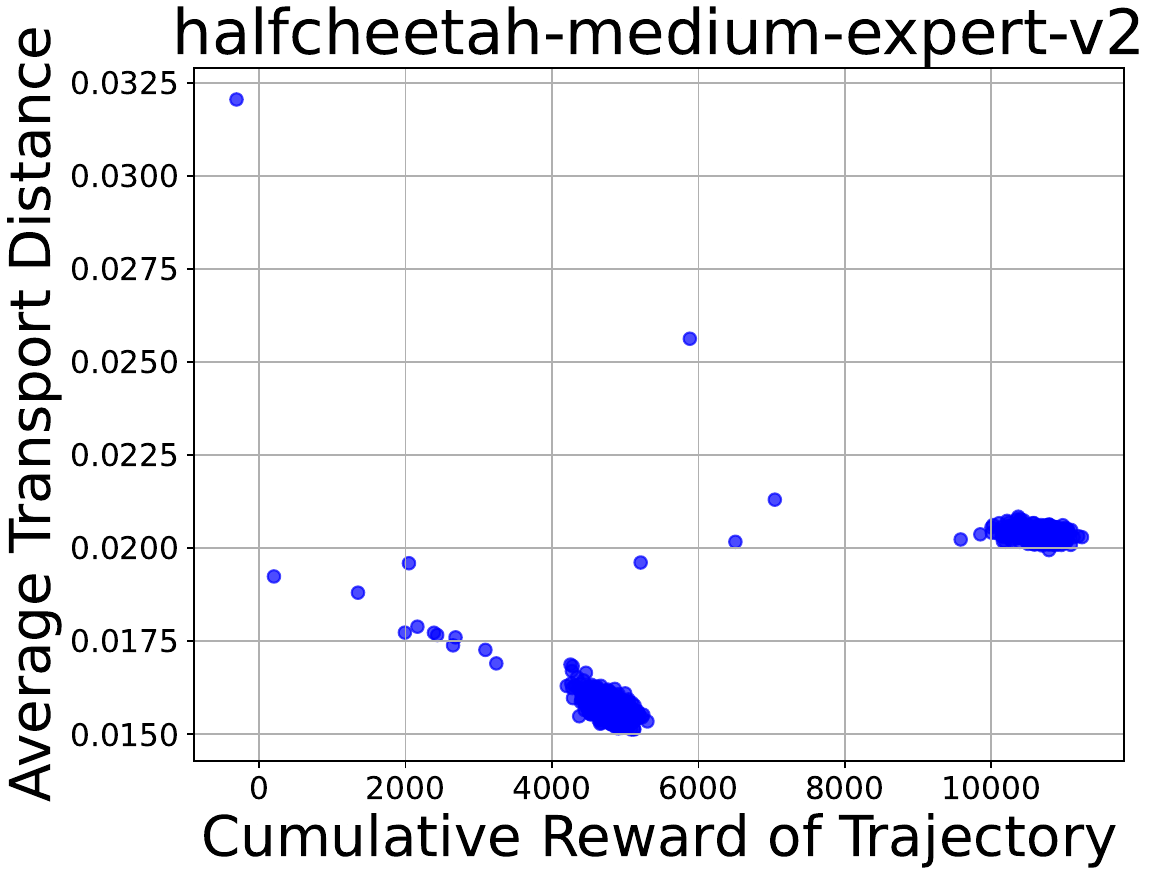}
    \end{minipage}
    \begin{minipage}{0.32\textwidth}
        \centering
        \includegraphics[width=\linewidth]{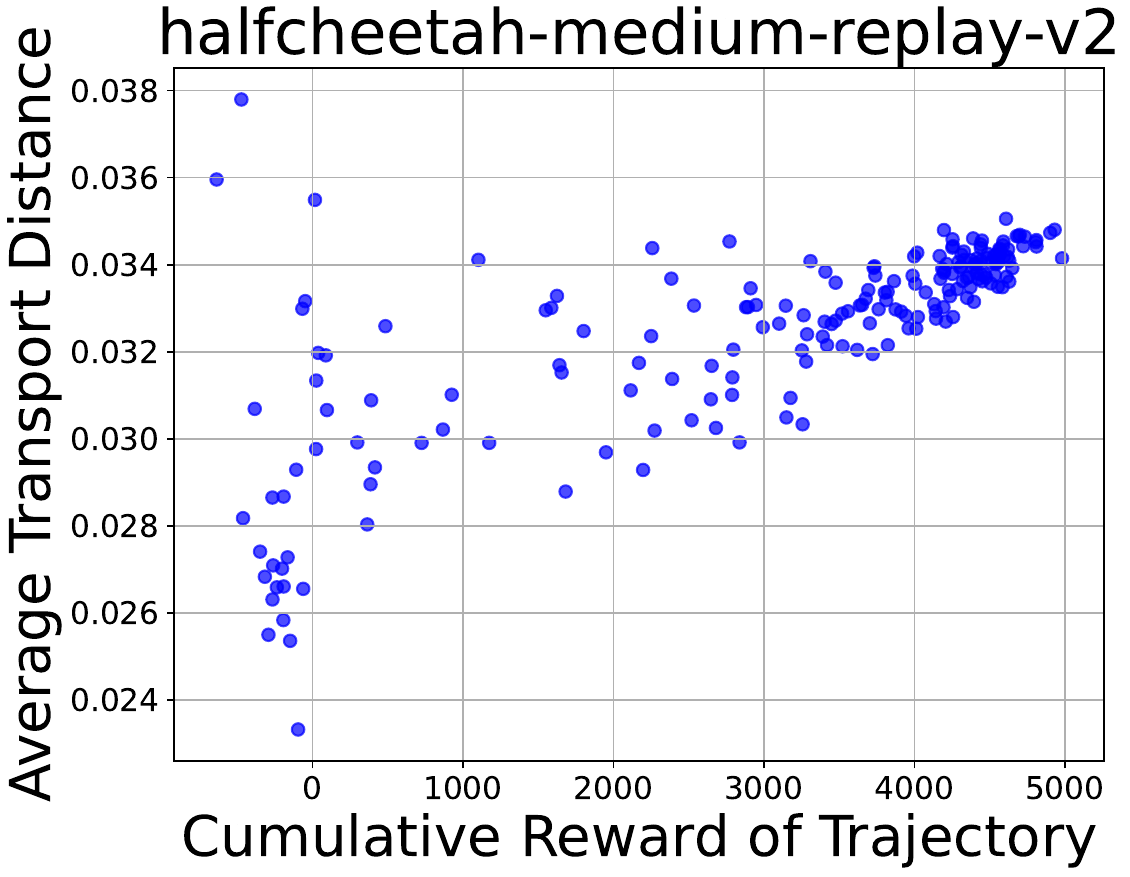}
    \end{minipage}
    
    \caption{The relationship between trajectory quality and transport-induced distance.}
    \label{fig:app_ret_dist}
\end{figure}

The results of other locomotion task are shown in \cref{fig:app_ret_dist}. In the Walker2d environment, similar to the Hopper environment, the transport distance was larger for lower-quality trajectories. In contrast, this tendency was not as clearly observed in the HalfCheetah environment. A smaller transport distance indicates that the transport that increases the advantage is not being identified by the value function. Thus, learning a value function capable of effectively transforming low-quality trajectories remains a challenge for future research in such tasks.

\end{document}